\pdfoutput=1

\documentclass[11pt]{article}

 \usepackage{emnlp2021}

\usepackage{times}
\usepackage{latexsym}

\usepackage[T1]{fontenc}

\usepackage[utf8]{inputenc}

\usepackage{microtype}

\usepackage{multirow}
\usepackage{amsmath}
\usepackage{amsfonts}
\usepackage{amssymb}
\usepackage{hyperref}
\hypersetup{
    colorlinks=true,
    linkcolor=blue
}

\usepackage{booktabs}

\usepackage{paralist}

\usepackage{mdwlist}
\usepackage{graphicx}

\usepackage{color}

\usepackage{bbm}

\usepackage[ruled,linesnumbered]{algorithm2e}

\usepackage{xspace}

\newcommand{\method}{KSTER\xspace}

\title{Learning Kernel-Smoothed Machine Translation with Retrieved Examples}

\author{Qingnan Jiang\textsuperscript{1}\thanks{~~Work is done while at ByteDance.}, Mingxuan Wang\textsuperscript{2}, Jun Cao\textsuperscript{2}, Shanbo Cheng\textsuperscript{2}, Shujian Huang\textsuperscript{1}\thanks{~~Corresponding author.}, Lei Li\textsuperscript{3}\footnotemark[1] \\
  \textsuperscript{1}National Key Laboratory for Novel Software Technology, Nanjing University, China \\
  \textsuperscript{2}ByteDance AI Lab, ~~~~~~~~~~~~~~~\textsuperscript{3}University of California, Santa Barbara \\
   \textsuperscript{1}\texttt{jiangqn@smail.nju.edu.cn, huangsj@nju.edu.cn} \\
   \textsuperscript{2}\texttt{\{wangmingxuan.89, caojun.sh, chengshanbo\}@bytedance.com} \\
   \textsuperscript{3}\texttt{lilei@cs.ucsb.edu}
}

\date{}

\begin{document}
\maketitle

\begin{abstract}

How to effectively adapt neural machine translation (NMT) models according to emerging cases without retraining?
Despite the great success of neural machine translation, updating the deployed models online remains a challenge.
Existing non-parametric approaches that retrieve similar examples from a database to guide the translation process are promising but are prone to overfit the retrieved examples.
In this work, we propose to learn Kernel-Smoothed Translation with Example Retrieval (\method), an effective approach to adapt neural machine translation models online.
Experiments on domain adaptation and multi-domain machine translation datasets show that even without expensive retraining, \method is able to achieve improvement of 1.1 to 1.5 BLEU scores over the best existing online adaptation methods.
The code and trained models are released at \href{https://github.com/jiangqn/KSTER}{https://github.com/jiangqn/KSTER}.

\end{abstract}

\section{Introduction}
\label{sec:intro}

Over the past years, end-to-end Neural Machine Translation (NMT) models have achieved great success~\cite{bahdanau2015neural,wu2016google,vaswani2017attention}. How to effectively update a deployed NMT model and adapt to emerging cases? For example, after a generic NMT model trained on WMT data, a customer wants to use service to translate financial documents.
The costomer may have a handful of translation pairs for the finance domain, but do not have the capacity to perform a full retraining.

Non-parametric adaptation methods enable incorporating individual examples on-the-fly, by retrieving similar source-target pairs from an external database to guide the translation process~\cite{bapna2019non,gu2018search,zhang2018guiding,cao2018encoding}.
The external database can be easily updated online.
Most of these methods rely on effective sentence-level retrieval.
Different from sentence retrieval, $k$-nearest-neighbour machine translation introduces token level retrieval to improve translation~\cite{khandelwal2021nearest}.
It shows promising results for online domain adaptation.

\begin{figure}
    \centering
    \includegraphics[width=7.8cm]{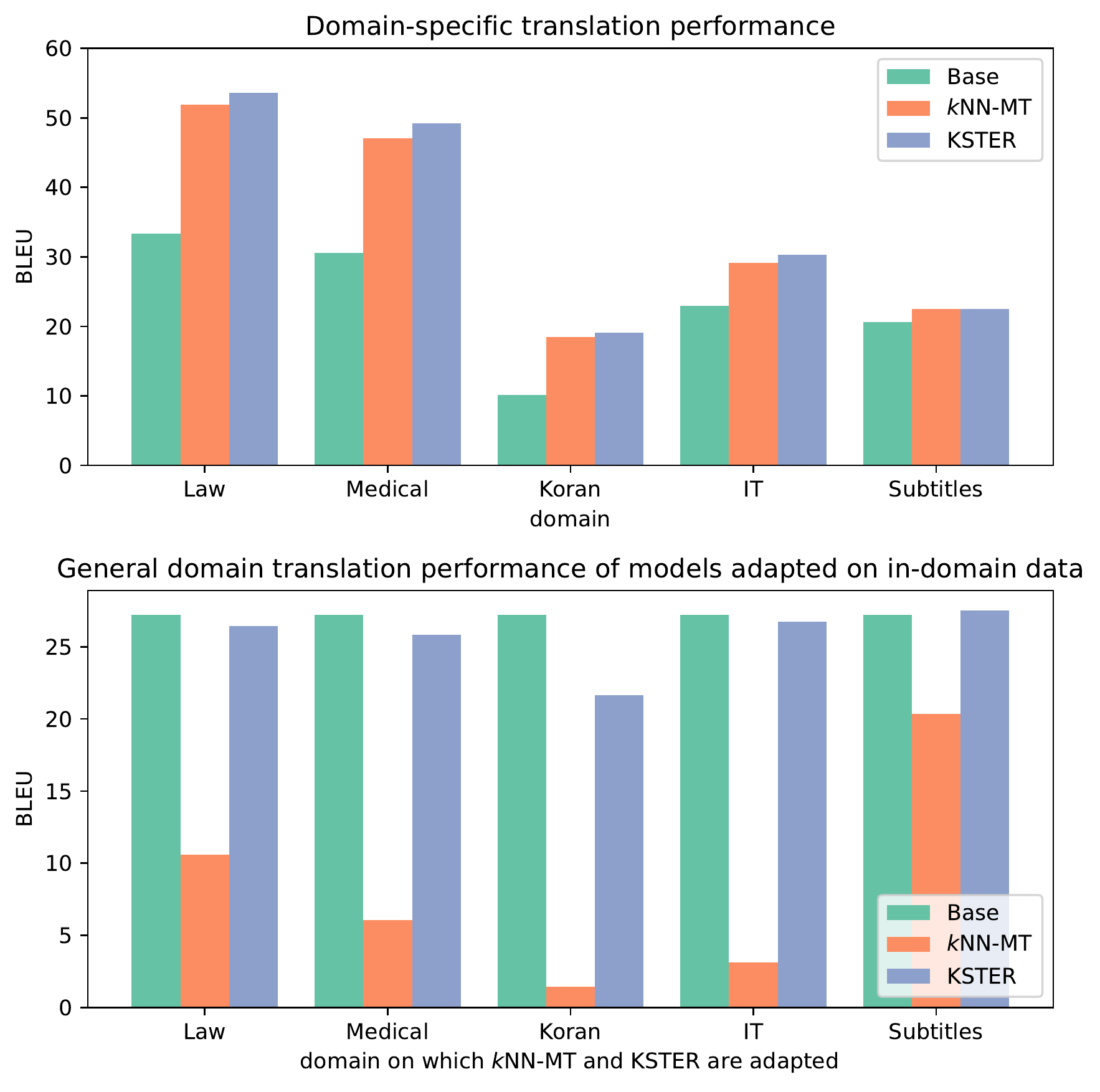}
    \caption{The domain-specific and general domain translation performance in EN-DE translation.
   Base is a Transformer model trained on general domain WMT data.
   $k$NN-MT and our proposed \method are adapted for domain-specific translation with in-domain database.
   Both $k$NN-MT and \method achieve improvements over Base in domain-specific translation performance.
   But $k$NN-MT overfits to in-domain data and performs bad in general domain translation, while the proposed \method achieves comparable general domain translation performance with Base.
   }
    \label{fig:overfit}
\end{figure}

There are still limitations for existing non-parametric methods for online adaptation.
First, since it is not easy for sentence-level retrieval to find examples that are similar enough to the test example,
this low overlap between test examples and retrieved examples brings noise to translation~\cite{bapna2019non}.
Second, completely non-parametric methods are prone to overfit the retrieved examples.
For example, although $k$NN-MT improves domain-specific translation, it overfits severely and can not generalize to the general domain, as is shown in Figure \ref{fig:overfit}.
An ideal online adaptation method should introduce less noise to the translation process and generalize to the changeful test examples with the incrementally changing database.

In this paper, we propose to learn Kernel-Smoothed Translation with Example Retrieval (\method), to effectively learn and adapt neural machine translation online.
\method retains the online adaptation advantage of non-parametric methods and avoids the drawback of easy overfitting.
More specifically, \method improves the generalization ability of non-parametric methods in three aspects. 
First, we introduce a learnable kernel to dynamically measure the relevance of the retrieved examples based on the current context. 
Then, the exampled-based distribution is combined with the model-based distribution computed by NMT with adaptive mixing weight for next word prediction.
Further, to make the learning of \method stable, we introduce a retrieval dropout strategy. The intuition is that similar examples can constantly be retrieved during training, but not the same situation during inference. 
We therefore drop the most similar examples during training to reduce this discrepancy.

With above improvements, \method shows the following advantages:
\begin{itemize}
    \vspace{-6pt}
    \setlength{\itemsep}{0pt}
    \setlength{\parsep}{0pt}
    \setlength{\parskip}{0pt}
    \item  Extensive experiments show that, \method outperforms $k$NN-MT, a strong competitor, in specific domains for 1.1 to 1.5 BLEU scores while keeping the performance in general domain.  
    \item  \method outperforms $k$NN-MT for 1.8 BLEU scores on average in unseen domains. Therefore, there is no strong restriction of the input domain, which makes \method much more practical for industry applications.  
    \vspace{-6pt}
\end{itemize}

\section{Related Work}
\label{sec:related}

This work is mostly related to two research areas in machine translation (MT), i.e., domain adaptation for machine translation and online adaptation of MT models by non-parametric retrieval.

\paragraph{Domain Adaptation for MT}~  Domain adaptation for MT aims to adapt general domain MT models for domain-specific language translation~\cite{chu2018a}.
The most popular method for this task is finetuning general domain MT models on in-domain data.
However, finetuning suffers from the notorious catastrophic forgetting problem~\cite{mccloskey1989catastrophic, santoro2016one}. 
There are also some sparse domain adaptation methods that only update part of the MT parameters~\cite{bapna2019simple, wuebker2018compact,liang2020finding, guo2021parameter,lin2021learning,zhu2021counter}.

In real-world translation applications, the domain labels of test examples are often not available.
This dilemma inspires a closely related research area --- multi-domain machine translation~\cite{pham2021revisiting, farajian2017multi, liang2020finding}, where one model translates sentences from all domains.

\paragraph{Online Adaptation of MT by Non-parametric Retrieval}

\noindent Non-parametric methods enable online adaptation of deployed NMT models by updating the database from which similar examples are retrieved.

Traditional non-parametric methods search sentence-level examples to guide the translation process~\cite{cao2018encoding, gu2018search, zhang2018guiding}.
Recently, n-gram level~\cite{bapna2019non} and token level~\cite{khandelwal2021nearest} retrieval are introduced and shows strong empirical results.
Generally, similar examples are retrieved based on fuzzy matching~\cite{bulte2019neural, xu2020boosting}, embedding similarity, or a mixture of the two approaches~\cite{bapna2019non}.
There are also works that utilize off-the-shelf search engine to retrieve similar translation examples~\cite{he2021fast, xia2019graph}.

\section{Methodology}
\label{sec:approach}
\begin{figure}
    \centering
    \includegraphics[width=7.8cm]{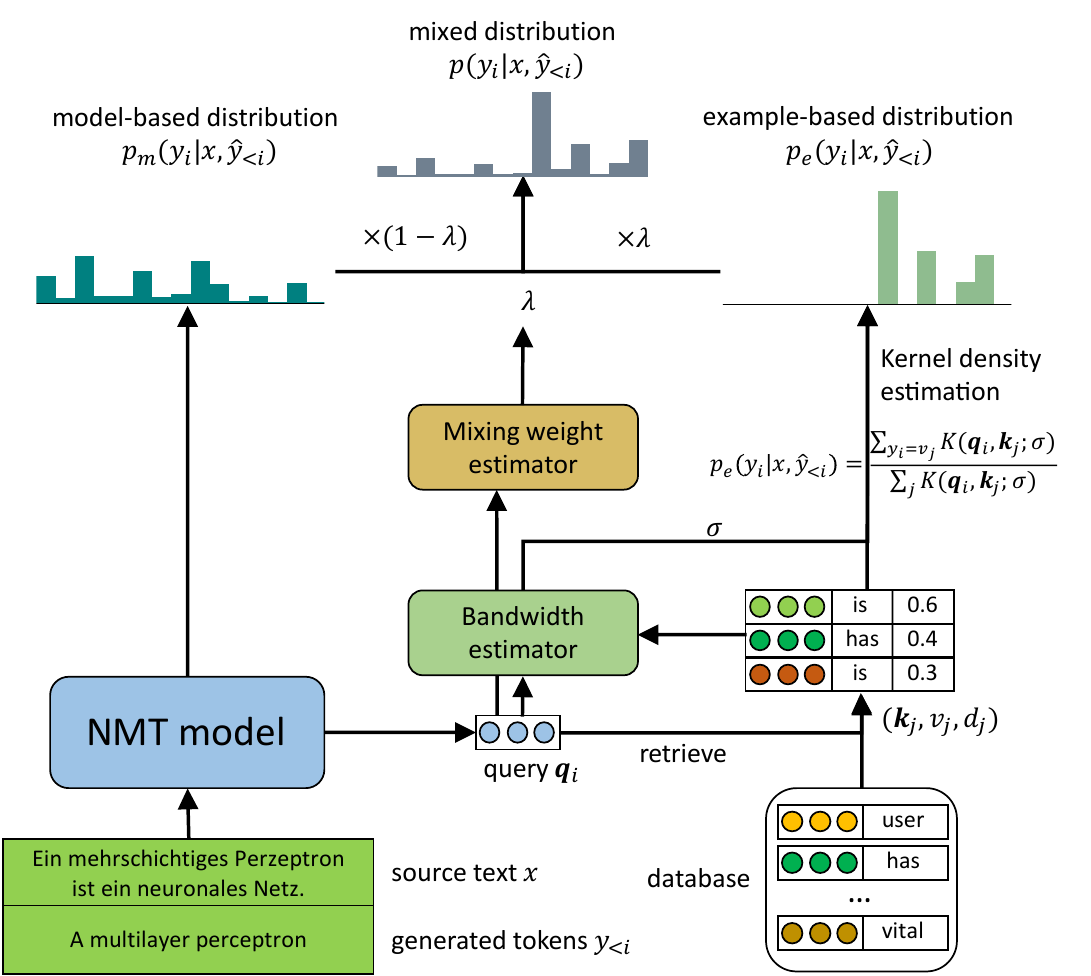}
    \caption{Overview of the \method. The left part is the NMT model denoted as base model. The right part shows the process of retrieving similar examples and estimating the example-based distribution.}
    \label{fig:overview}
\end{figure}

In this section, we first formulate the kernel-smoothed machine translation (\method), which smooths neural machine translation (NMT) output with retrieved token level examples.
Then we introduce the modeling and training of the learnable kernel and adaptive mixing weight.
The overview of \method is shown in Figure \ref{fig:overview}.

\subsection{Kernel-Smoothed Machine Translation}

\paragraph{Base Model for Neural Machine Translation}

The state-of-the-art NMT models are based on the encoder-decoder architecture.
The encoder encodes the source text $x$ into a sequence of hidden states.
The decoder takes the representations of the source text as input and generates target text auto-regressively. 
In each decoding step, the decoder predicts the probability distribution of next tokens $p_{m}(y_i|x, \hat{y}_{<i}; \theta)$ conditioned on the source text $x$ and previously generated target tokens $\hat{y}_{<i}$. The NMT model parameters are denoted as $\theta$.

\paragraph{Kernel Smoothing with Retrieved Examples}

The model-based distribution $p_{m}(y_i|x, \hat{y}_{<i}; \theta)$ is then smoothed by an example-based distribution $p_{e}(y_i|x, \hat{y}_{<i})$. It is computed using kernel density estimation (KDE) on retrieved examples.

We build a database from which similar examples are retrieved.
The database consists of all token level examples from the training set in the form of key-value pairs.
In each key-value pair $(\mathbf{k}, v)$, the key $\mathbf{k} = f_{\text{NMT}}(x, y_{<i}; \theta)$ is the intermediate representation of a certain layer in the NMT decoder.
The value is the corresponding ground truth target token $y_i$.
The key $\mathbf{k}$ can be seen as a vector representation of the context of value $v$.
We obtain the key-value pairs from $(x, y)$ by running force-decoding with a trained NMT model.

In each decoding step, we compute the query $\mathbf{q}_i = f_{\text{NMT}}(x, \hat{y}_{<i}; \theta)$ and retrieve $k$ similar examples based on the $L^2$ distance \footnote{For two $d$-dimension vectors $\mathbf{x}$ and $\mathbf{y}$, we compute the $L^2$ distance between $\mathbf{x}$ and $\mathbf{y}$ as $\sqrt{\sum_{i=1}^{d}(\mathbf{x}_i - \mathbf{y}_i)^2}$.} query and keys.
Each retrieved example forms a triple $(\mathbf{k}_j, v_j, d_j)$, where $\mathbf{k}_j$ is the key; $v_j$ is the corresponding value token and $d_j$ is the $L^2$ distance between query $\mathbf{q}_i$ and key $\mathbf{k}_j$.
The example-based distribution $p_{e}(y_i|x, \hat{y}_{<i})$ is then computed with these retrieved examples using the following equation.

\begin{equation}
    p_{e}(y_i|x, \hat{y}_{<i}) = \frac{\sum_{y_i=v_j}K(\mathbf{q}_i, \mathbf{k}_j; \sigma)}{\sum_{j}K(\mathbf{q}_i, \mathbf{k}_j; \sigma)}
\end{equation}

where $K(\mathbf{q}, \mathbf{k}; \sigma)$ is the kernel function and $\sigma$ is the parameter of the kernel.

Finally, the NMT output is smoothed by combing the model-based distribution and the example-based distribution with a mixing weight $\lambda$.

\begin{align}
    p(y_i|x, \hat{y}_{<i}; \theta) & = \lambda p_{e}(y_i|x, \hat{y}_{<i}) \\
    & + (1 - \lambda) p_{m}(y_i|x, \hat{y}_{<i}; \theta)
\end{align}

\begin{figure}
    \centering
    \includegraphics[width=7.8cm]{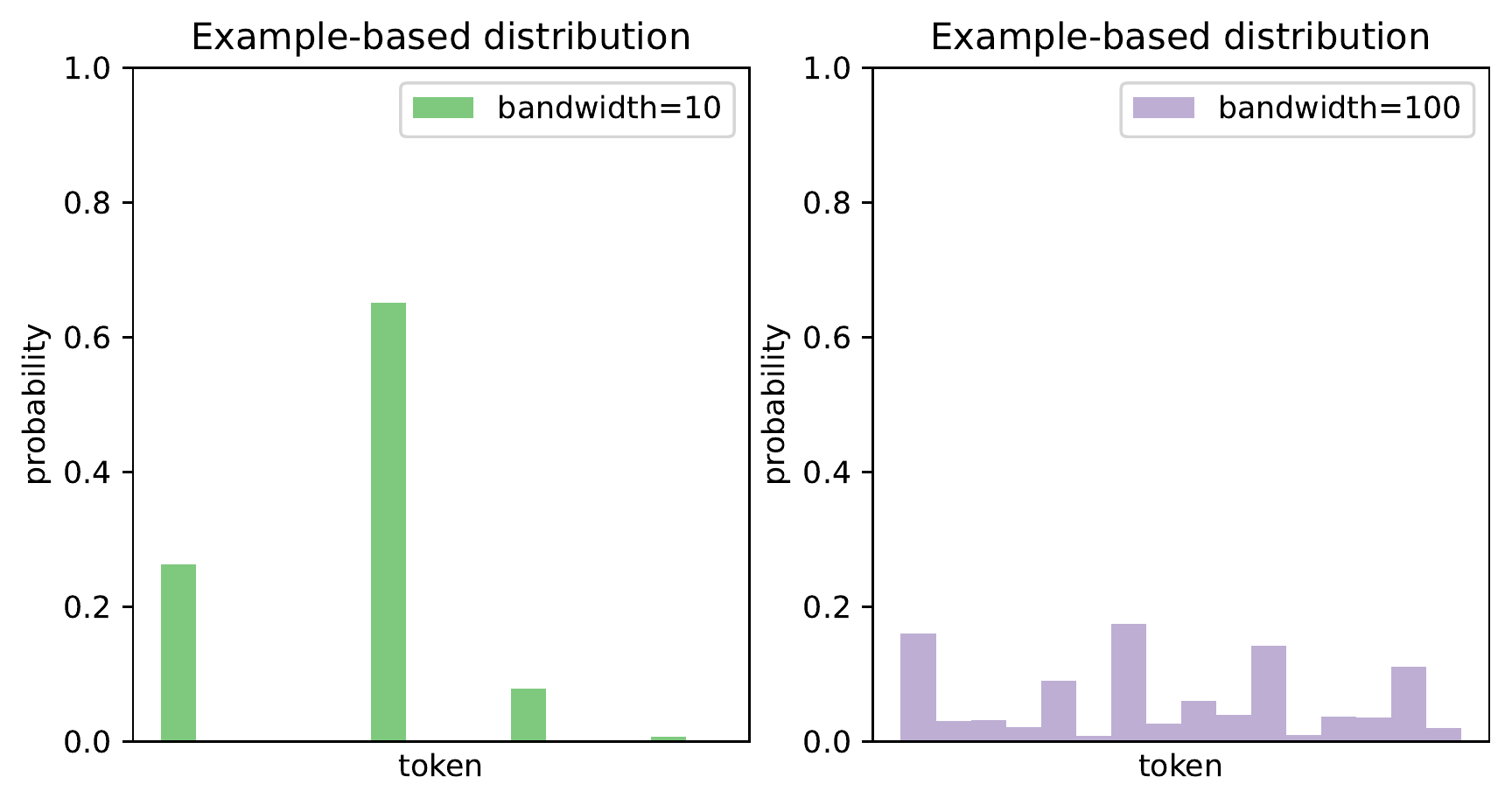}
    \caption{A visualization of example-based distribution estimated by Gaussian kernel KDE with different bandwidth. Bandwidth controls the smoothness of example-based distribution.}
    \label{fig:bandwidth}
\end{figure}

\subsection{Learnable Kernel Function}

Although all kernel functions can be used in KDE, we choose two specific kernels in this study --- Gaussian kernel and Laplacian kernel, since they are easy to parameterize.

The only parameter in Gaussian kernel $K_{g}(\mathbf{q}_i, \mathbf{k}_j; \sigma) = \exp(-\frac{\|\mathbf{q}_i - \mathbf{k}_j\|^2}{\sigma})$ is the bandwidth $\sigma$,
it controls the smoothness of the example-based distribution, as is shown in Figure \ref{fig:bandwidth}.

In a learnable Gaussian kernel, the bandwidth is not a fixed hyper-parameter.
Instead, it is estimated in each decoding step by a learned affine network with exponential activation.

\begin{equation}
    \mathbf{\sigma} = \exp(\mathbf{W}_{1}[\mathbf{q}_{i}; \overline{\mathbf{k}}_{i}] + \mathbf{b}_{1})
\end{equation}

 $\overline{\mathbf{k}}_{i} = \frac{1}{k}\sum_{j=1}^{k}\mathbf{k}_{j}$ is average-pooled keys and $[\mathbf{W}_1; \mathbf{b}_1]$ are trainable parameters.
 
The bandwidth of learnable Laplacian kernel $K_{l}(\mathbf{q}_i, \mathbf{k}_j; \sigma) = \exp(-\frac{\|\mathbf{q}_i - \mathbf{k}_j\|}{\sigma})$ is modeled in the same way as the bandwidth of learnable Gaussian kernel.

\subsection{Adaptive Mixing of Base Prediction and Retrieved Examples}

To mix the model-based distribution and example-based distribution adaptively, we model the mixing weight $\lambda$ with a learnable neural network.

The mixing weight $\lambda$ is computed by a multi-layer perceptron with query $\mathbf{q}_{i}$ and weighted sum of keys $\widetilde{\mathbf{k}}$ as inputs, where $[\mathbf{W}_2;\mathbf{b}_2; \mathbf{W}_3; \mathbf{b}_3]$ are trainable parameters.

\begin{gather}
    \mathbf{\lambda} = \text{sigmoid}(\mathbf{W}_{3}\text{ReLU}(\mathbf{W}_{2}[\mathbf{q}_{i}; \widetilde{\mathbf{k}}_{i}] + \mathbf{b}_{2}) + \mathbf{b}_3) \\
    \widetilde{\mathbf{k}}_{i}= \sum_{j=1}^{k}w_j\mathbf{k}_{j} \\
    w_j \propto K(\mathbf{q}_i, \mathbf{k}_j; \theta)
\end{gather}







In this way, $k$NN-MT~\cite{khandelwal2021nearest} could be seen as a specific case of \method, with fixed Gaussian kernel and mixing weight.

\begin{table}
    \small
    \centering
    \begin{tabular}{c|ccccc}
        \hline
         & Law & Medical & Koran & IT & Subtitles \\
        \hline
        Train & 467k & 248k & 18k & 223k & 500k \\
        Dev & 2k & 2k & 2k & 2k & 2k \\
        Test & 2k & 2k & 2k & 2k & 2k \\
        \hline
    \end{tabular}
    \caption{The number of training, development and test examples of 5 domain-specific datasets. The training data of Subtitles domain is sampled from the full Subtitles training set by \citet{aharoni2020unsupervised}.}
    \label{tab:domain-statistics}
\end{table}

\subsection{Training}

We optimize the \method model by minimizing the cross entropy loss between the mixed distribution and ground truth target tokens:

\begin{equation}
    \min_{\phi} -\sum_{i=1}^{n} \log p(y_i|x, y_{<i};\theta, \phi)
\end{equation}

where $n$ is the length of a target sentence $y$.

We keep the NMT model parameters $\theta$ fixed.
Only parameters of learnable kernel and mixing weight $\phi = [\mathbf{W}_1; \mathbf{b}_1; \mathbf{W}_2; \mathbf{b}_2; \mathbf{W}_3; \mathbf{b}_3]$ are trained.

\subsection{Retrieval Dropout}

Since the database is built from the training data and \method is trained on the training data, similar examples can constantly be retrieved from the database during training.
However, in test time, there may be no example in the database that is similar to the query.
This discrepancy between training and inference may lead to overfitting.

We propose a simple training strategy called retrieval dropout to alleviate this problem.
During training, we search top $k + 1$ similar examples instead of top $k$ examples. 
Then we drop the most similar example and use the remaining $k$ examples for training.
Retrieval dropout is not used in test time.

\section{Experiments}
\label{sec:exps}


We evaluate the proposed methods on two machine translation tasks: domain adaptation for machine translation (DAMT) and multi-domain machine translation (MDMT).
In DAMT, in-domain translation model is built for each specific domain, since the domain labels of examples are available in test time.
In MDMT, the domain labels of examples are not available in test time, so examples from all domains are translated with one model, which is a more practical setting.

\begin{table}
    \small
    \centering
    \begin{tabular}{c|ccccc}
        \hline
         & Law & Medical & Koran & IT & Subtitles \\
        \hline
        EN-DE & 16.1M & 6.5M & 0.5M & 3.4M & 6.3M \\
        DE-EN & 15.8M & 6.2M & 0.5M & 3.0M & 6.4M \\
        \hline
    \end{tabular}
    \caption{The database size --- number of examples, of each domain in DAMT.}
    \label{tab:database}
\end{table}

\begin{table*}
\small
    \centering
    \begin{tabular}{c|c|ccccc|cc}
        \hline
         Direction & Methods & Law & Medical & Koran & IT & Subtitles & Average-specific & Average-general (WMT14) \\
        \hline
        \multirow{5}{*}{EN-DE}
        & Base & 33.36 & 30.54 & 10.16 & 22.99 & 20.65 & 23.54 & \textbf{27.20} \\
        & Finetuning & 49.07 & 47.10 & \textbf{25.98} & \textbf{36.28} & \textbf{26.00} & \textbf{36.89} & 14.17 \\
        & $k$NN-MT & 51.88 & 47.02 & 18.51 & 29.12 & 22.46 & 33.80 & 8.32 \\
        & \method & \textbf{53.63} & \textbf{49.18} & 19.10 & 30.28 & 22.54 & 34.95 & 25.63 \\
        \hline
        \multirow{5}{*}{DE-EN} 
        & Base & 36.80 & 33.36 & 11.24 & 29.21 & 23.13 & 26.75 & \textbf{31.49} \\
        & Finetuning & 55.19 & 51.35 & \textbf{22.87} & \textbf{41.88} & \textbf{28.33} & \textbf{39.92} & 17.82 \\
        & $k$NN-MT & 57.40 & 50.92 & 15.74 & 34.92 & 25.38 & 36.87 & 13.18 \\
        & \method & \textbf{59.41} & \textbf{53.40} & 16.97 & 35.74 & 25.94 & 38.29 & 30.23 \\
        \hline
    \end{tabular}
    \caption{Test set BLEU scores of DAMT. Laplacian kernel is used in \method. Average-specific and average-general domain represent the averaged performance of adapted models in domain-specific translation and general domain translation. \method outperforms $k$NN-MT for 1.2 and 1.4 BLEU scores on average in EN-DE and DE-EN directions. Significance test by paired bootstrap resampling shows that \method outperforms $k$NN-MT significantly in all domains except for Subtitles domain in EN-DE translation and IT domain in DE-EN translation.}
    \label{tab:domain-adaptation}
\end{table*}

\subsection{Datasets and Implementation Details}

\paragraph{Datasets} 
We conduct experiments in EN-DE translation and DE-EN translation.
We use WMT14 EN-DE dataset~\cite{bojar2014findings} as general domain training data, which consists of 4.5M sentence pairs.
\textit{newstest2013} and \textit{newstest2014} are used as the general domain development set and test set, respectively.
5 domain-specific datasets proposed by \citet{koehn2017six} and re-splited by \citet{aharoni2020unsupervised}\footnote{\href{https://github.com/roeeaharoni/unsupervised-domain-clusters}{https://github.com/roeeaharoni/unsupervised-domain-clusters}} are used to evaluate the domain-specific translation performance.
The detailed statistics of the 5 datasets are shown in Table \ref{tab:domain-statistics}.


\paragraph{Implementation Details} 
We use joint Byte Pair Encoding (BPE)~\cite{sennrich2016neural} with 30k merge operations for subword segmentation.
The resulted vocabulary is shared between source and target languages.
We employ Transformer Base~\cite{vaswani2017attention} as the base model.
Following \citet{khandelwal2021nearest}, the normalized inputs of feed forward network in the last Transformer decoder block are used as keys to build the databases and queries for retrieval.
The translation performance is evaluated with detokenized BLEU scores \cite{papineni2002bleu}, computed by SacreBLEU~\cite{post2018a} \footnote{\href{https://github.com/mjpost/sacrebleu}{https://github.com/mjpost/sacrebleu}}.

We build a FAISS~\cite{Johnson17Billion} index for nearest neighbour search.
We employ inverted file and product quantization to accelerate retrieval in large scale databases.
The keys of examples are stored in the fp16 format to reduce the memory demand.
We set $k = 16$ to keep a balance between translation quality and inference speed.

We train the base model for 200k steps.
The best 5 checkpoints are averaged to obtain the final model.
We train \method for 30k steps.
For the training procedures of all models, 
each batch contains 32,768 tokens approximately.
The models are optimized by Adam optimizer \cite{kingma2015adam} with learning rates set to 0.0002.

\method introduced 526k trainable parameters, which is 0.85\% of the base model.
We implement all the models based on JoeyNMT \cite{kreutzer2019joey} \footnote{\href{https://github.com/joeynmt/joeynmt}{https://github.com/joeynmt/joeynmt}}. 

\subsection{Domain Adaptation for Machine Translation}

We build individual database for each specific domain with in-domain training data in DAMT.
The sizes of databases are shown in Table \ref{tab:database}.

\paragraph{Baselines}
We compare the proposed method with the following baselines.

\begin{itemize}
    \vspace{-6pt}
    \setlength{\itemsep}{0pt}
    \setlength{\parsep}{0pt}
    \setlength{\parskip}{0pt}
    \item \textbf{Base} base model trained on general-domain data.
    \item \textbf{Finetuning} base model trained on general domain dataset and then finetuned with in-domain data for each specific domain individually.
    \item \textbf{$k$NN-MT} $k$NN-MT with in-domain database individually, where the hyper-parameters are tuned on development set of each domain.
    \vspace{-6pt}
\end{itemize}

The \method model is trained for each specific domain individually for fair comparison . 

\begin{table}
    \small
    \centering
    \begin{tabular}{c|cc}
        \hline
        Method & EN-DE & DE-EN \\
        \hline
        $k$NN-MT & 33.80 & 36.87 \\
        + $10\%$ source noise & 31.26 (-2.54) & / \\
        + $10\%$ target noise & / & 33.43 (-3.44) \\
        \hline
        \method & 34.78 & 38.17 \\
        + $10\%$ source noise & 32.89 (-1.89) & / \\
        + $10\%$ target noise & / & 35.67 (-2.50) \\
        \hline
    \end{tabular}
    \caption{The averaged BLEU scores over 5 specific domains in DAMT with noisy database. \method is more robust than $k$NN-MT when the quality of database is not good.}
    \label{tab:robustness}
\end{table}

\paragraph{Main results}
The DAMT experiment results are shown in Table \ref{tab:domain-adaptation}.
For domain-specific translation, \method outperforms $k$NN-MT for 1.2 and 1.4 BLEU scores on average in EN-DE and DE-EN translation respectively.
Finetuning achieves best domain-specific performance on average. 
However, the performance of finetuned models on general domain drops significantly due to the catastrophic forgetting problem.
The even worse general domain performance of $k$NN-MT indicates that it overfits to the retrieved examples severely.
\method performs far better than finetuning and $k$NN-MT in general domain, which shows strong generalization ability.
We notice that \method with Laplacian kernel performs slightly better than Gaussian kernel, since \method with Gaussian kernel tends to ignore the long-tailed retrieved examples.

\begin{table*}
\small
    \centering
    \begin{tabular}{c|c|c|ccccc|c}
        \hline
         Direction & Methods & General (WMT14) & Law & Medical & Koran & IT & Subtitles & Average-specific \\
        \hline
        \multirow{5}{*}{EN-DE}
        & Base & 27.20 & 33.36 & 30.54 & 10.16 & 22.99 & 20.65 & 23.54 \\
        & Joint-training & 27.25 & 45.02 & 44.52 & 15.43 & \textbf{34.48} & \textbf{25.16} & 32.92 \\
        & $k$NN-MT & 24.72 & 51.24 & 46.54 & \textbf{16.29} & 29.55 & 21.80 & 33.08 \\
        & \method & \textbf{27.69} & \textbf{53.04} & \textbf{49.23} & 15.94 & 31.82 & 22.63 & \textbf{34.53} \\
        \hline
        \multirow{5}{*}{DE-EN} 
        & Base & 31.49 & 36.80 & 33.36 & 11.24 & 29.21 & 23.13 & 26.75 \\
        & Joint-training & 31.62 & 50.95 & 47.48 & \textbf{18.13} & \textbf{39.57} & \textbf{27.73} & 36.77 \\
        & $k$NN-MT & 25.87 & 57.38 & 50.83 & 14.57 & 37.56 & 22.86 & 36.64 \\
        & \method & \textbf{31.94} & \textbf{58.64} & \textbf{52.79} & 15.24 & 36.90 & 25.15 & \textbf{37.74} \\
        \hline
    \end{tabular}
    \caption{Test set BLEU scores of multi-domain machine translation. Average-specific is the averaged performance in 5 specific domains. For general domain sentence translation, \method outperforms $k$NN-MT for 3 and 6 BLEU scores in EN-DE and DE-EN direction respectively.
For domain-specific translation, \method outperforms $k$NN-MT for 1.5 and 1.1 BLEU scores in EN-DE and DE-EN direction. Significance test by paired bootstrap~\cite{koehn2004statistical} resampling shows that \method outperforms $k$NN-MT significantly in all domains except for Koran domain in EN-DE translation and IT domain in DE-EN translation.}
    \label{tab:multi-domain}
\end{table*}

\paragraph{Robustness test}
The performance of MT model with non-parametric retrieval is influenced by the size and quality of database.
\citet{khandelwal2021nearest} have studied how translation performance of $k$NN-MT changes with the size of database.
In this work, we study the performance change of $k$NN-MT and \method with low-quality database.
Specifically, we test the robustness of these models in DAMT when the database is noisy.

We add token-level noise to the English sentences in parallel training data by EDA~\cite{wei2019eda} \footnote{We do not experiment with adding noise to the German side, since German WordNet is not available for us, which is necessary for synonym replacement}.
For each word in a sentence, it is modified with a probability of $0.1$.
The candidate modifications contain synonym replacement, random insertion, random swap and random deletion with equal probability.
Then we use the noisy training data to construct the noisy database.

We study the effects of source side noise and target side noise on translation performance.
The experiment results are presented in Table \ref{tab:robustness}.
Target side noise has more negative effect to translation performance than source side noise.
The BLEU scores of \method drop less apparently in all settings, which indicates that the proposed method is more robust with low-quality database.

\subsection{Multi-Domain Machine Translation}

In MDMT, since there is no domain label available in test time, examples from all domains are translated with one model. 
We build a mixed database with training data of general domain and 5 specific domains, which is used in all MDMT experiments.
The mixed database for EN-DE translation and DE-EN translation contains 172M and 167M key-value pairs respectively.

\paragraph{Baselines}

We compare the proposed method with the following baselines.

\begin{itemize}
    \vspace{-6pt}
    \setlength{\itemsep}{0pt}
    \setlength{\parsep}{0pt}
    \setlength{\parskip}{0pt}
    \item \textbf{Base} base model trained on general domain dataset.
    \item \textbf{Joint-training} base model trained on the mixture of general domain dataset and 5 specific domain datasets.
    \item \textbf{$k$NN-MT} $k$NN-MT with mixed database. The hyper-parameters are selected that achieve highest averaged development set BLEU scores over general domain and 5 specific domains.
    \vspace{-6pt}
\end{itemize}

\begin{figure}
    \centering
    \includegraphics[width=7.8cm]{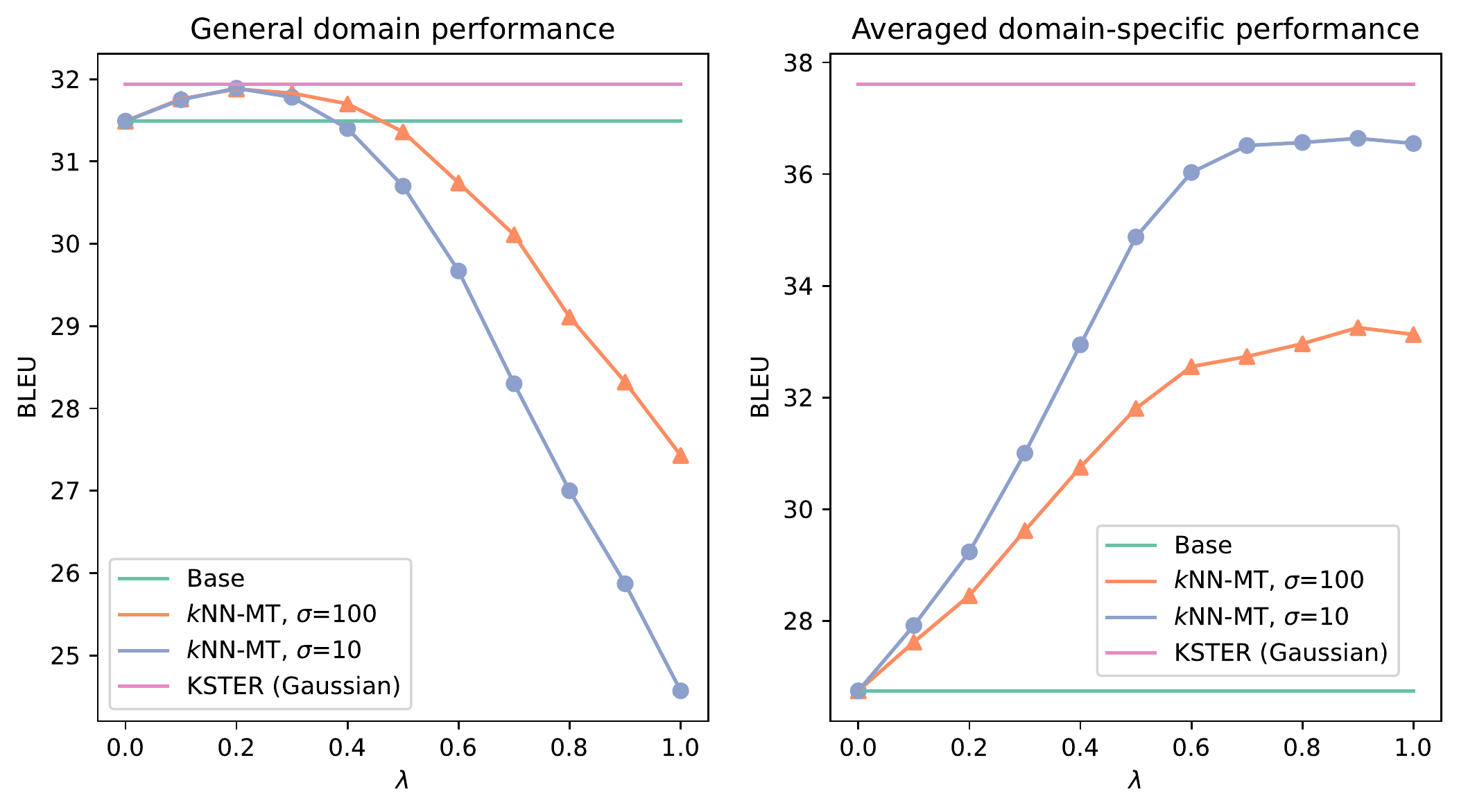}
    \caption{General domain and averaged domain-specific performance of $k$NN-MT with different hyper-parameter selections in DE-EN direction, together with the performance of Base and \method with Gaussian kernel.}
    \label{fig:weight-temperature-effects}
\end{figure}

We sample 500k training examples from general domain training set, which are then mixed with all 5 specific domain training examples for \method training.

\paragraph{Main results}
The experiment results of MDMT are shown in Table \ref{tab:multi-domain}.
For general domain sentence translation, \method outperforms $k$NN-MT for 3 and 6 BLEU scores in EN-DE and DE-EN direction respectively.
For domain-specific translation, \method outperforms $k$NN-MT for 1.5 and 1.1 BLEU scores in EN-DE and DE-EN direction.
Besides, \method also outperforms joint-training in both general domain performance and averaged domain-specific performance significantly.
The proposed method achieves advantages over joint-training in both online adaptation and translation performance.

\begin{figure}
    \centering
    \includegraphics[width=7.8cm]{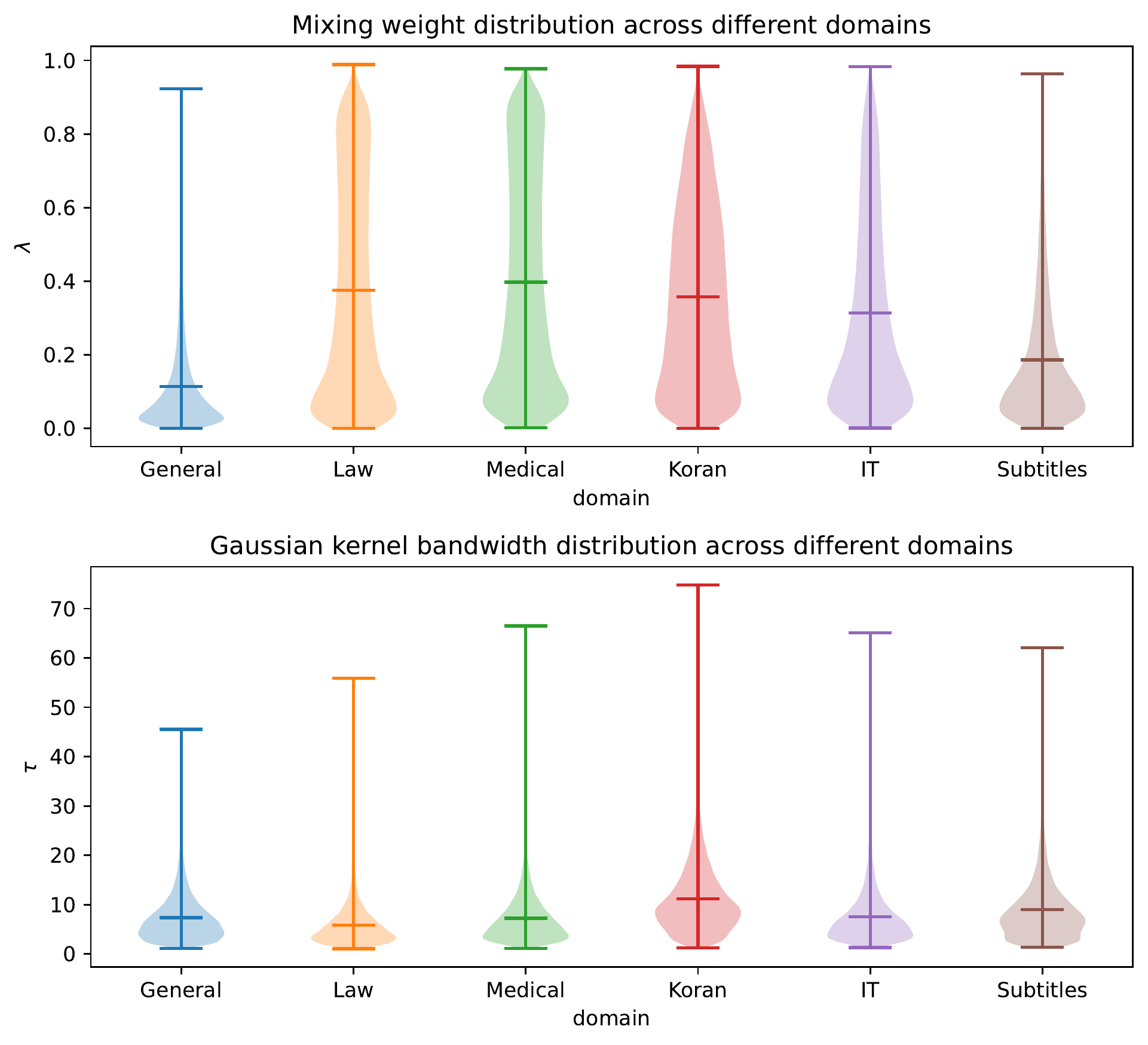}
    \caption{The bandwidth and mixing weight distribution of \method with Gaussian kernel.}
    \label{fig:knn-weight}
\end{figure}

\paragraph{General-specific domain performance trade-off}

We plot the general domain performance and averaged domain-specific performance of $k$NN-MT with different hyper-parameter selections in Figure \ref{fig:weight-temperature-effects}.
We find that $k$NN-MT performs better in domain-specific translation when the system prediction relies more on the searched examples (low bandwidth and higher mixing weight).
In contrast, better general domain translation performance is achieved when the system prediction relies more on NMT prediction (high bandwidth and low mixing weight).

There is a trade-off between general and specific domain performance in $k$NN-MT.
Applying the identical kernel and mixing weight for all test examples can not achieve best performance in general domain and specific domains simultaneously.

\method with Gaussian kernel, which is a generalization of $k$NN-MT, achieves better performance in both general domain and domain-specific translation since it applies adaptive kernel and mixing weight for different test examples.
Distributions in Figure \ref{fig:knn-weight} indicates that \method learns different kernels and different weights for different examples.

\paragraph{Generalization ability over unseen domains}

We test the generalization ability of baselines and \method with Laplacian kernel in unseen domains, which is important in real-world MDMT applications. 
We take Bible and QED from OPUS~\cite{tiedemann2012parallel}\footnote{\href{https://opus.nlpl.eu/}{https://opus.nlpl.eu/}} as unseen domains and randomly sample 2k examples from each domain for test.
We directly use the MDMT models to translate sentences from unseen domains.
The results of EN-DE translation are presented in Table \ref{tab:unseen-domains}.
\method outperforms all baselines, which shows strong generalization ability.

\begin{table}
\small
    \centering
    \begin{tabular}{c|cc|c}
        \hline
        Method & Bible & QED & Averaged \\
        \hline
        Base & 12.51 & 25.21 & 18.86 \\
        Joint-training & 12.69 & 25.90 & 19.30 \\
        $k$NN-MT & 12.35 & 23.56 & 17.96 \\
        \method (Laplacian) & \textbf{13.32} & \textbf{26.16} & \textbf{19.74} \\
        \hline
    \end{tabular}
    \caption{The test set BLEU scores of MDMT models in unseen domains.
    \method generalizes better than all baselines in unseen domains.}
    \label{tab:unseen-domains}
\end{table}

\subsection{Inference Speed}

A common concern about non-parametric methods in MT is that searching similar examples may slow the inference speed.
We test the inference speed \method in MDMT in EN-DE translation, which is the setting with the largest database. 
The averaged inference time in general domain and 5 specific domain test sets of $k$NN-MT is 1.15 times of the base model.
The averaged inference time of \method is 1.19 times of the base model, which is only slightly slower than the baselines.

\section{Analysis}
\label{sec:analysis}
\begin{table}
\small
    \centering
    \begin{tabular}{c|cc|cc}
        \hline
         \multirow{2}{*}{} & \multicolumn{2}{c|}{EN-DE} & \multicolumn{2}{c}{DE-EN}  \\
         \cline{2-5}
         & General & Specific & General & Specific \\
        \hline
         None & 24.72 & 33.08 & 25.87 & 36.64 \\
         Kernel & 26.06 & 33.40 & 27.89 & 37.37 \\
         Weight & \textbf{27.80} & 34.02 & 31.88 & 37.19 \\
         Both & 27.74 & \textbf{34.38} & \textbf{31.94} & \textbf{37.61} \\
        \hline
    \end{tabular}
    \caption{Ablation study of learnable kernel and mixing weight in \method with Gaussian kernel in MDMT. Both learnable kernel and learnable mixing weight bring improvement. None represents that both kernel and mixing weight are fixed, in which case \method degenerates to $k$NN-MT.}
    \label{tab:parameterization-ablation}
\end{table}

In this section, we first conduct ablation studies to verify the effectiveness of each part of the proposed method.
Then we conduct detailed analysis on how kernel-smoothing with retrieved examples helps translation.

\subsection{Ablation Studies of Proposed Methods}

\paragraph{Both learnable kernel and learnable mixing weight bring improvement}

In \method, both the kernel and mixing weight are learnable.
We study the effect of keeping only one of the two parts learnable in MDMT.

We take \method with Gaussian kernel for analysis. 
The ablation experiment results are presented in Table \ref{tab:parameterization-ablation}.

Both learnable kernel and learnable mixing weight bring improvement in both general domain and domain-specific translation.
Keeping the two parts learnable simultaneously brings additional improvement.
Overall, learnable mixing weight is more important than learnable kernel function.

\paragraph{\method outperforms $k$NN-MT with all $k$ selections}

\begin{figure}
    \centering
    \includegraphics[width=7.8cm]{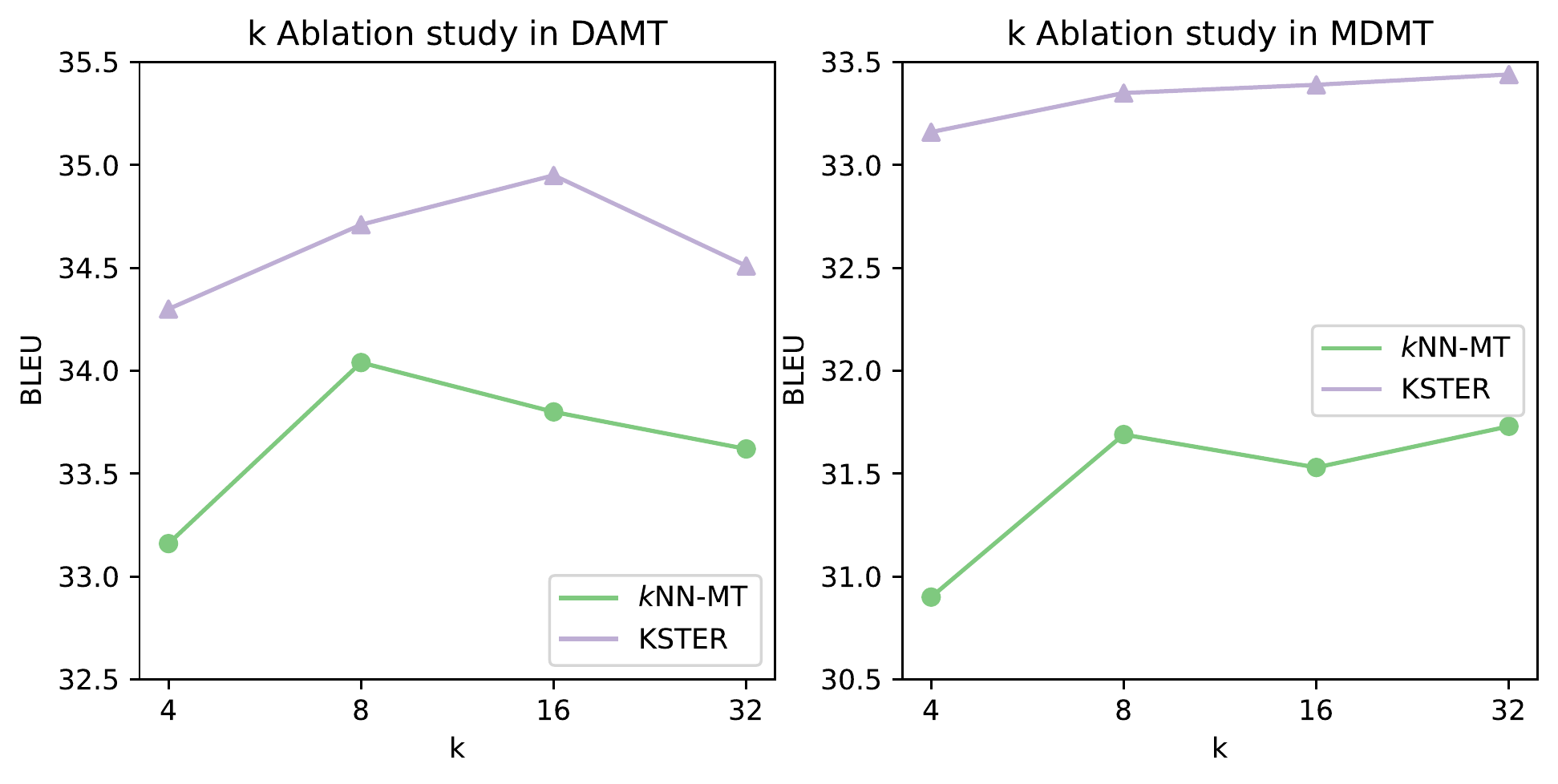}
    \caption{The ablation study of $k$ selections in DAMT and MDMT settings in EN-DE translation. In DAMT, we report the averaged performance on 5 specific domains. In MDMT, we report the averaged performance on general domain and 5 specific domains. With all $k$ selections, \method outperforms $k$NN-MT consistently.}
    \label{fig:k-ablation}
\end{figure}

We conduct ablation study over different $k$ selections in both DAMT and MDMT settings in EN-DE translation.
We experiment with four $k$ selections --- \{4, 8, 16, 32\}, and plot the results in Figure \ref{fig:k-ablation}.
In DAMT, \method achieves best performance with $k = 16$.
In MDMT, the performance of our method increases with $k$.
With all $k$ selections, \method outperforms $k$NN-MT consistently.

\paragraph{Retrieval dropout improves generalization}

We study the effect of retrieval dropout in MDMT and select the \method with Laplacian kernel for analysis.

\begin{figure}
    \centering
    \includegraphics[width=7.8cm]{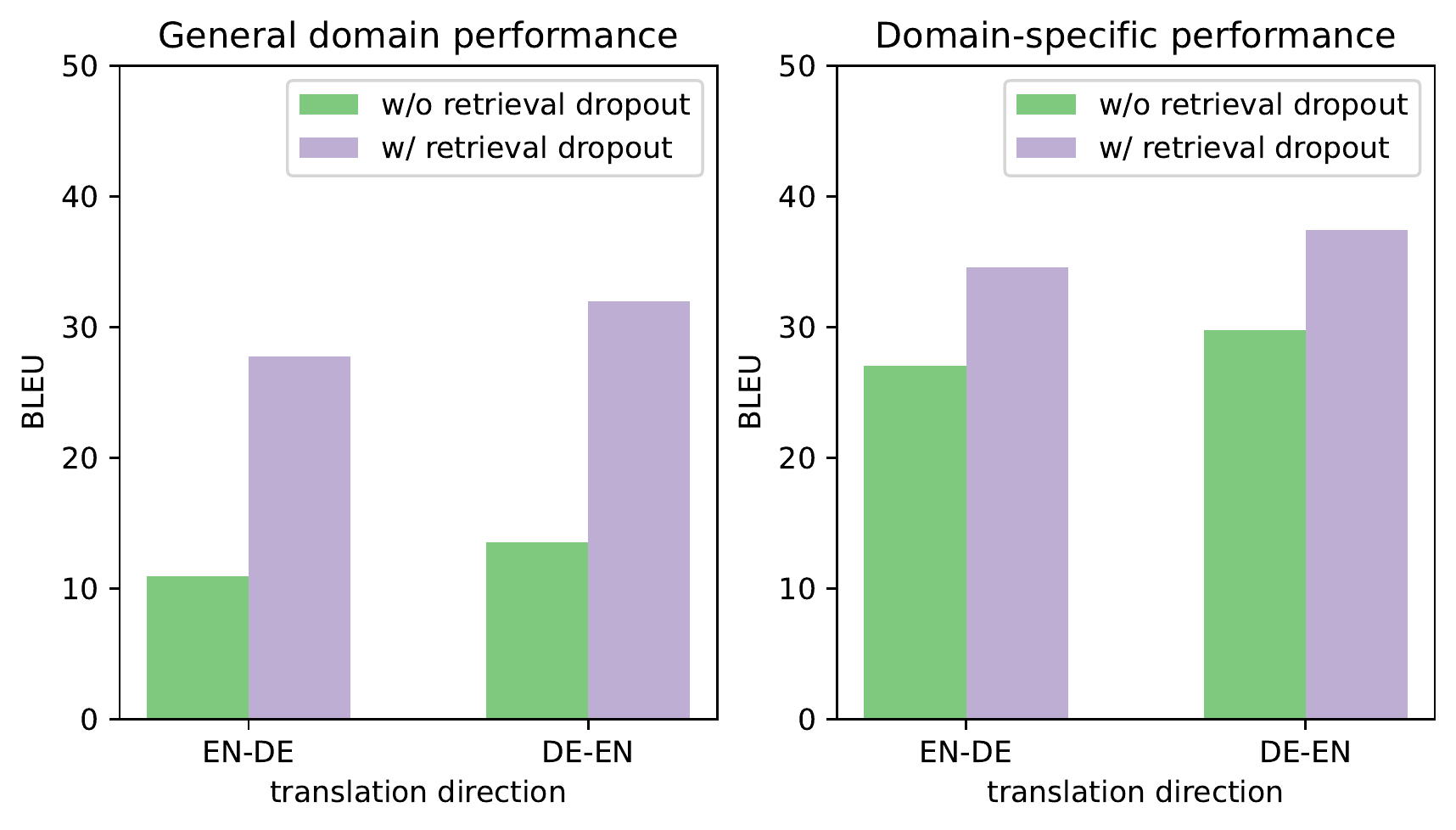}
    \caption{The ablation study of \method with Laplacian kernel in MDMT. Without retrieval dropout, \method overfits severely.}
    \label{fig:knn-dropout-ablation}
\end{figure}

We plot the general domain and averaged domain-specific translation performance of models trained with or without retrieval dropout in Figure \ref{fig:knn-dropout-ablation}.
Without retrieval dropout, the performance of both general domain and domain-specific translation drops dramatically.
The discrepancy between training and inference leads to severe overfitting.
This problem is alleviated by the proposed retrieval dropout, which shows that this training strategy improves the generalization ability of \method.

\subsection{Fine-grained Effects of Kernel-smoothing with Retrieved Examples on Translation}

For better understanding the effects of kernel-smoothing with retrieved examples on translation, we study the following two research questions.

\begin{itemize}
    \vspace{-6pt}
    \setlength{\itemsep}{0pt}
    \setlength{\parsep}{0pt}
    \setlength{\parskip}{0pt}
    \item [-] RQ1. Which types of word kernel-smoothing influences most?
    \item [-] RQ2. Does kernel-smoothing help word sense disambiguation?
    \vspace{-6pt}
\end{itemize}

\paragraph{Kernel-smoothing influences verbs, adverbs and nouns most}

To study the first research question, we categorize the predicted words with their Part-of-Speech tags (POS tags).
In each decoding step, if the predicted word $y_i$ obtains the highest probability of example-based distribution but it does not obtain the highest probability of model-based distribution, it is recognized as a prediction determined by kernel-smoothing with retrieved examples.

We compute the ratio of predictions determined by kernel-smoothing across different POS tags.
We conduct this analysis on DAMT task in EN-DE direction and select Medical and Subtitles domains as representatives.

\begin{figure}
    \centering
    \includegraphics[width=7.8cm]{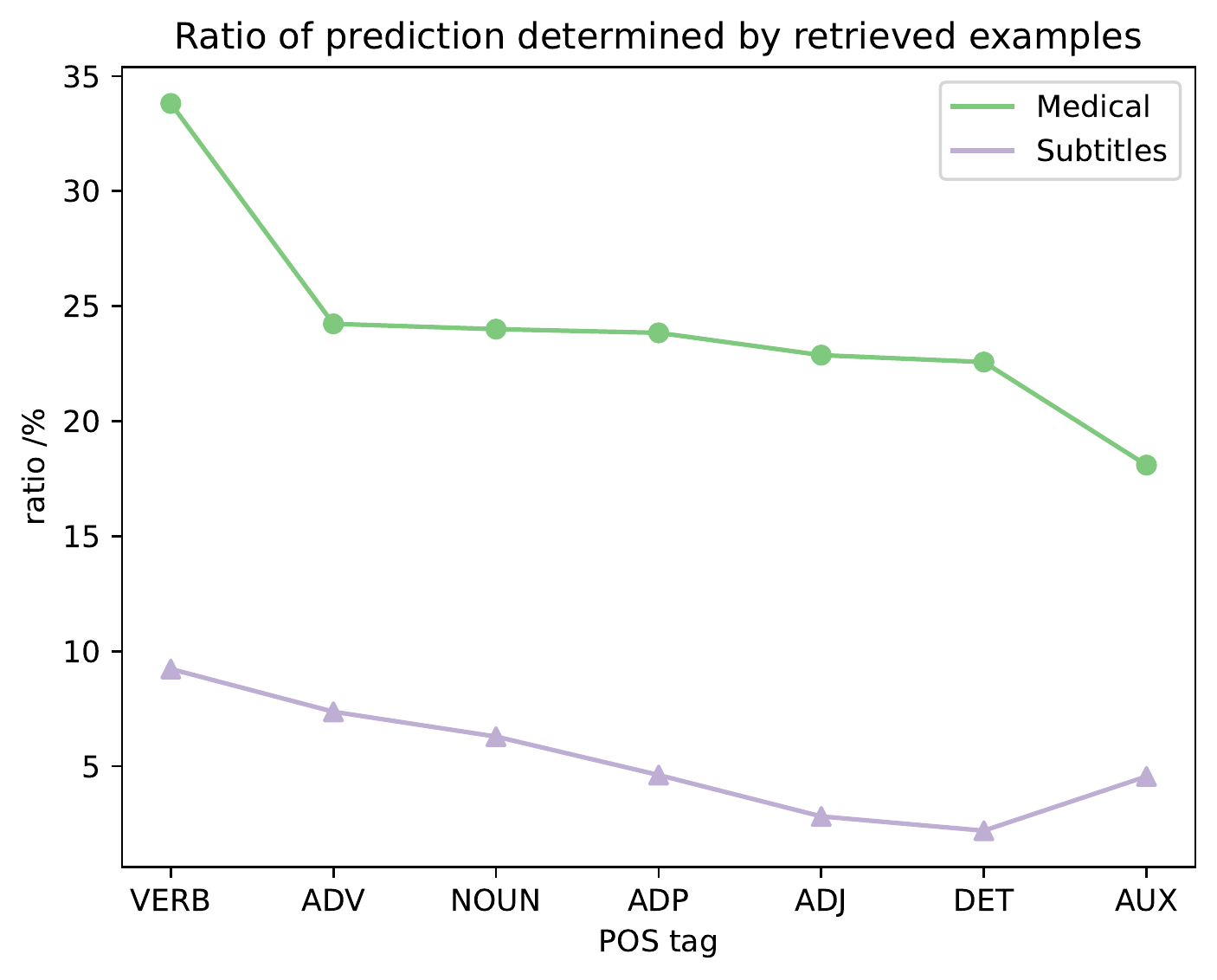}
    \caption{The ratio of word predictions determined by kernel-smoothing. Kernel-smoothing influences verbs, adverbs and nouns most.}
    \label{fig:pos-tag-analysis}
\end{figure}

We report the results in Figure \ref{fig:pos-tag-analysis}.
Medical and Subtitles represent two opposite cases where non-parametric retrieval contributes more in the former and contributes less in the latter.
We find that across the 2 different domains, kernel-smoothing contributes most to the predictions of verbs, adverbs and nouns, which are morphologically complex word types.
Retrieving words in similar context may helps selecting the correct form of morphologically complex words.

\paragraph{Kernel-smoothing helps word sense disambiguation}

In kernel-smoothing, we search examples with similar keys --- contextualized hidden states.
We hypothesize that the retrieved examples contains useful context information which helps word sense disambiguation.
We test this hypothesis with contrastive translation pairs.

A contrastive translation pair contains a source, a reference and one or more contrastive translations.
Contrastive translations are constructed by replacing a word in reference with a word which is another translation of an ambiguous word in the source.
NMT systems are used to score reference and contrastive translations.
If an NMT system assign higher score to reference than all contrastive translations in an example, the NMT system is recognized as making correct prediction on this example.

We use ContraWSD~\cite{gonzales2017improving} \footnote{\href{https://github.com/ZurichNLP/ContraWSD}{https://github.com/ZurichNLP/ContraWSD}} as the test suite, which contains 7,359 contrastive translation pairs for DE-EN translation.
We encode the source sentences from ContraWSD and training data of 5 specific domains by averaged BERT embeddings~\cite{devlin2018bert}.
Then we whiten the sentence embeddings with BERT-whitening proposed by \citet{huang2021whiteningbert, li2020on}.
For each domain, we select 300 examples from ContraWSD that most similar to the in-domain data based on the cosine similarity of sentence embeddings.

\begin{figure}
    \centering
    \includegraphics[width=7.8cm]{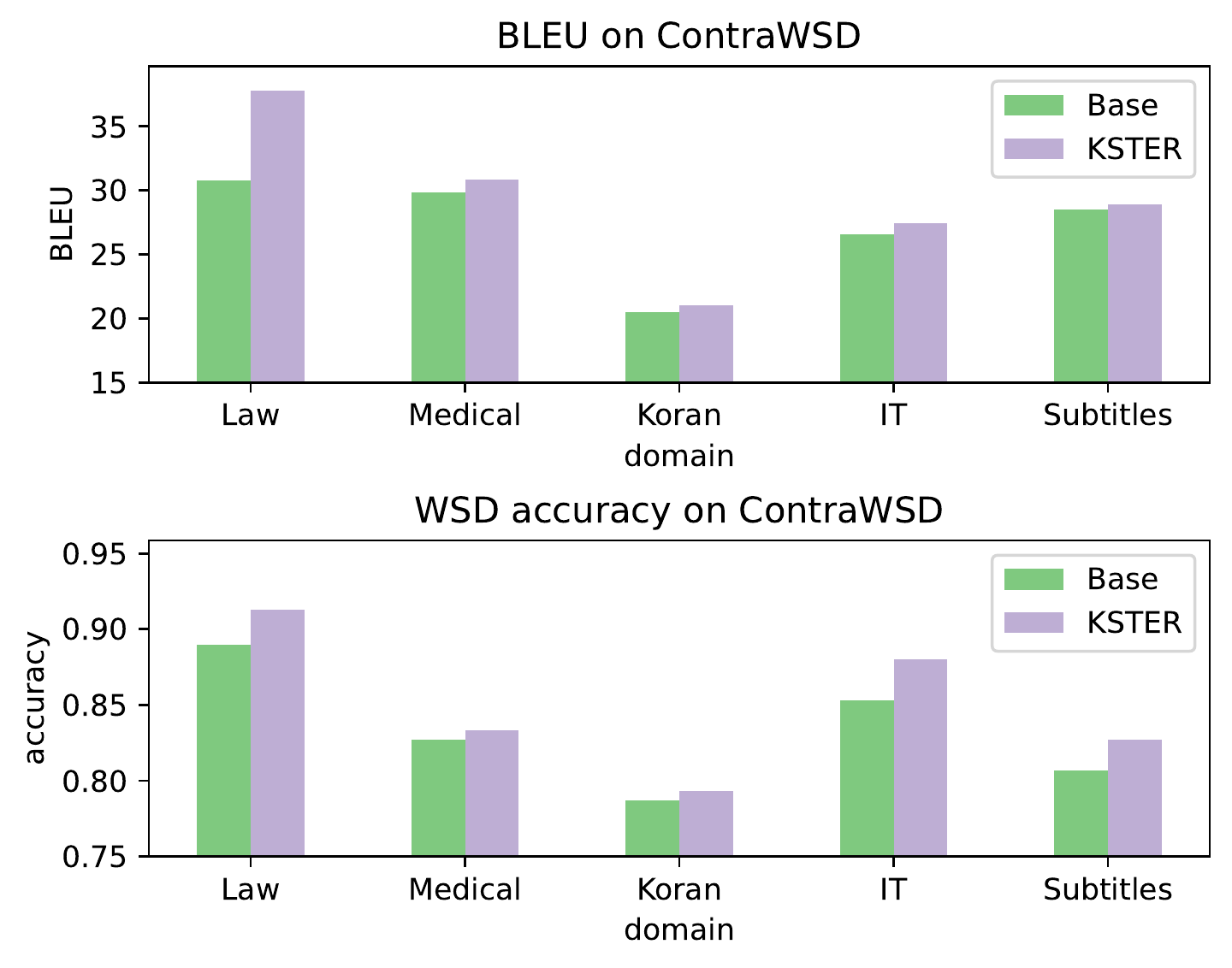}
    \caption{BLEU and word sense disambiguation accuracy of base model and \method with Gaussian kernel on ContraWSD dataset. Kernel-smoothing helps word sense disambiguation.}
    \label{fig:wsd}
\end{figure}

We evaluate the translation performance and word sense disambiguation ability of base model and \method for MDMT on selected examples for each domain.
The results are shown in Figure \ref{fig:wsd}.
Experimental results show that \method consistently outperforms base model in both translation performance and word sense disambiguation accuracy, which indicates that kernel-smoothing helps word sense disambiguation in machine translation.

\section{Conclusion}
\label{sec:conclusion}
In this work, we propose kernel-smoothed machine translation with retrieved examples.
It improves the generalization ability over existing non-parametric methods, while keeps the advantage of online adaptation.

\section{Acknowledgements}
\label{sec:acknowledgements}
We would like to thank the anonymous reviewers for their insightful comments. This work is supported by National Science Foundation of China (No. U1836221, 61772261),  National Key R\&D Program of China (No. 2019QY1806).

\bibliographystyle{acl_natbib}
\bibliography{paper}

\end{document}